%% file: main.tex
\documentclass[runningheads]{llncs}

\usepackage{graphicx}
\usepackage{booktabs}

\usepackage[accsupp]{axessibility}  

\usepackage{multirow}

\usepackage{hyperref}

\usepackage{orcidlink}

\usepackage{siunitx}

\usepackage{xcolor}

\definecolor{minyuecolor}{HTML}{21A8AC}

\newcommand{\pms}[1]{\ensuremath{\raisebox{0.5ex}{\scriptsize$\scriptscriptstyle\pm\!#1$}}}

\usepackage{wrapfig}

\begin{document}

\title{Controllable Text-to-Motion Generation via Modular Body-Part Phase Control} 

\titlerunning{Controllable T2M via Body-Part Phase Control}


\author{
Minyue Dai\inst{1} \and
Ke Fan\inst{2} \and
Anyi Rao\inst{3} \and
Jingbo Wang\inst{4} \and
Bo Dai\inst{5}
}

\authorrunning{M.~Dai et al.} 

\institute{
$^{1}$Fudan University \quad
$^{2}$Shanghai Jiao Tong University \quad
$^{3}$HKUST \quad
$^{4}$Shanghai AI Laboratory \quad
$^{5}$The University of Hong Kong
}

\maketitle

\begin{figure}[h] 
    \vspace{-2em}
    \centering
    \includegraphics[width=0.7\textwidth]{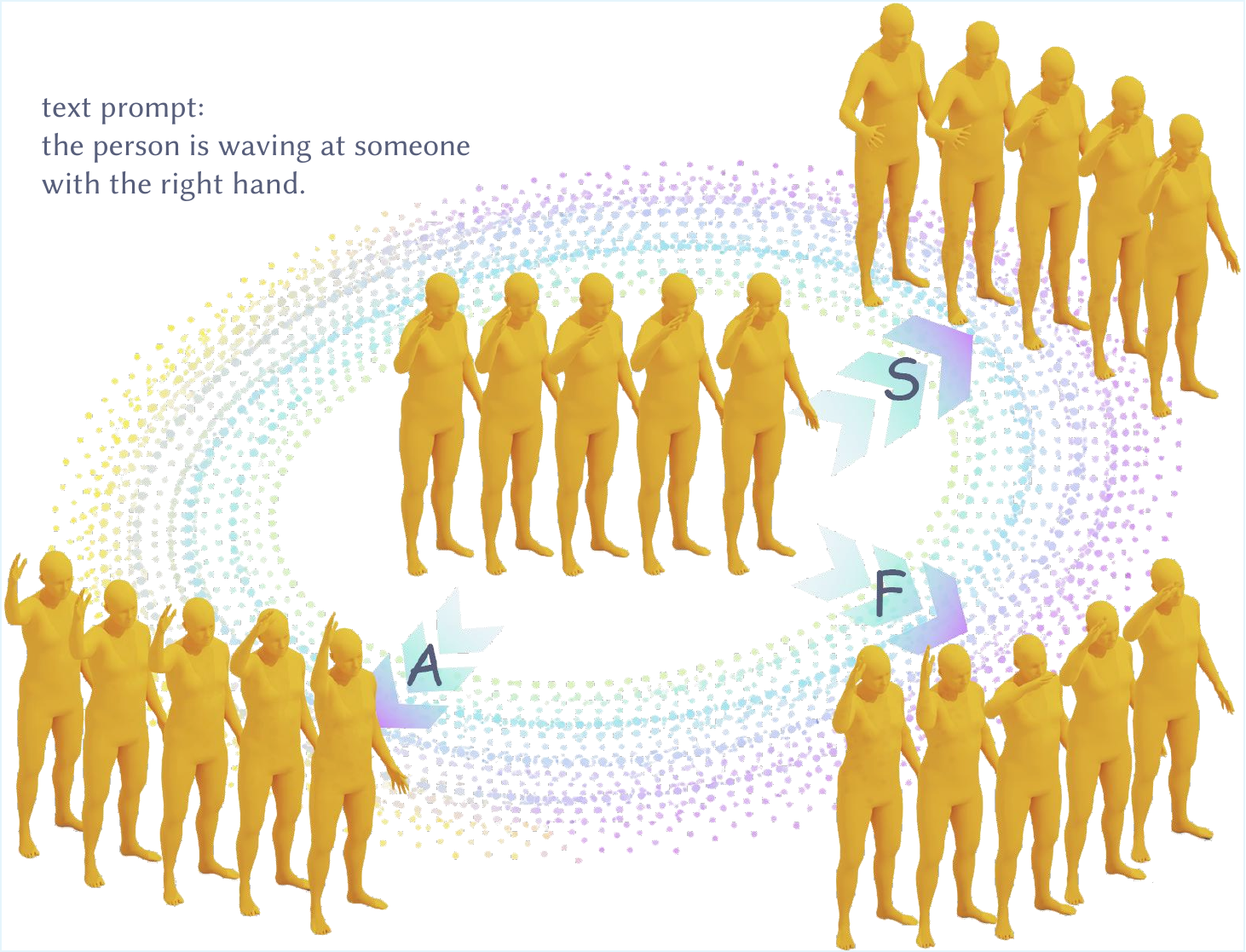}
    \vspace{-0.5em}
    \caption{
    Our method enables localized motion control via body part phase.
    By scalar editing the phase parameters of a target body part, namely amplitude (A), frequency (F), and phase shift (S), we can directly modulate its motion magnitude, repetition pace, and temporal alignment in the generated sequence.
%
    }
    \vspace{-2.5em}
    \label{fig:teaser}
\end{figure}

\input{sec/0.abstract}
\input{sec/1_introduction_fk}

\input{sec/2.related_work}
\input{sec/3.preliminaries}
\input{sec/4_method_fk}
\input{sec/5.experiments}
\input{sec/6.conclusion}







\bibliographystyle{splncs04}

\bibliography{main}

\end{document}

%% file: sec/0.abstract.tex
\begin{abstract}

Text-to-motion (T2M) generation is becoming a practical tool for animation and interactive avatars. 
However, modifying specific body parts while maintaining overall motion coherence remains challenging.
Existing methods typically rely on cumbersome, high-dimensional joint constraints (e.g., trajectories), which hinder user-friendly, iterative refinement. 
To address this, we propose \textbf{Modular Body-Part Phase Control}, a plug-and-play framework enabling structured, localized editing via a compact, scalar-based phase interface. 
By modeling body-part latent motion channels as sinusoidal phase signals—characterized by amplitude, frequency, phase shift, and offset—we extract interpretable codes that capture part-specific dynamics. 
A modular Phase ControlNet branch then injects this signal via residual feature modulation, seamlessly decoupling control from the generative backbone. 
Experiments on both diffusion- and flow-based models demonstrate that our approach provides predictable and fine-grained control over motion magnitude, speed, and timing. 
It preserves global motion coherence and offers a practical paradigm for controllable T2M generation.
Project page: \url{https://jixiii.github.io/bp-phase-project-page/}.

\end{abstract}

%% file: sec/1_introduction_fk.tex
\section{Introduction}
\label{sec:introduction}

Recent advances in text-to-motion (T2M) ~\cite{mdm,t2mgpt,zhou2023let,jiang2023motiongpt,momask,fan2025go} generation have enabled the synthesis of realistic and semantically aligned human motions from natural language descriptions. 
These approaches increasingly utilize diffusion-based and flow-based generative models, which have demonstrated strong performance in capturing complex motion distributions and producing highly diverse results. 
However, most existing approaches focus primarily on global motion generation, offering limited mechanisms for fine-grained, localized control. 
In practical applications such as character animation and virtual avatar manipulation, users often require structured, body-part-level adjustments (e.g., enlarging arm swings, changing leg stepping pace, or shifting gesture timing) while keeping the remaining body parts coherent and largely unchanged. 
Despite rapid progress, it still remains a fundamental challenge to enable such predictable and localized control in text-to-motion generation.

Prior works~\cite{mdm,xie2023omnicontrol,dai2024motionlcm,karunratanakul2023guided,pinyoanuntapong2024mmm,wan2024tlcontrol,guo2025motionlab} explored controllability in diffusion-based T2M generation, where joint-trajectory and keyframe constraints are incorporated through mechanisms such as condition injection, inpainting, or ControlNet-based branches. Representative approaches, such as OmniControl~\cite{xie2023omnicontrol} and MotionLCM~\cite{dai2024motionlcm}, integrate spatial-temporal control signals or trajectory encoders to condition generation on specific control joints. While effective for enforcing explicit geometric targets, such controls are often cumbersome for interactive use. They typically require constructing dense, per-frame joint-level specifications; consequently, small, intuitive edit requests (e.g., ``slightly earlier'' or ``twice as large'') do not easily translate into simple adjustments of the control signal. Alternatively, text-based motion editing methods~\cite{athanasiou2024motionfix,petrovich2024multi,li2025simmotionedit} rely on text prompts to guide motion modifications from a source to a target. 
However, these methods typically require large-scale, costly paired datasets, and often suffer from imprecise control due to the inherent ambiguity of natural language. 
Therefore, there is a critical need for a balanced approach that achieves flexible, fine-grained control while allowing for precise and mathematically predictable edits.

To support intuitive small edits and enable predictable, localized control, we propose \textbf{Modular Body-Part Phase Control}, a plug-and-play framework with an interpretable body-part control interface. 
Our key insight is that kinematic phase\footnote{In biomechanics and signal processing, a \textit{phase} representation characterizes the state of a periodic or quasi-periodic motion cycle. 
It parameterizes movement through distinct attributes such as amplitude (magnitude), frequency (pace), and phase shift (timing).} serves as an excellent descriptor for human motion dynamics. 
By parameterizing motion into frequency, amplitude, and phase shift, we can effectively capture and manipulate the speed, magnitude, and rhythm of human movements. 
To our knowledge, this is the first attempt to introduce a structured phase representation for fine-grained body-part control in generative motion models. 

Specifically, our framework comprises three core components. First, a time-dependent phase manifold is formed by a \textbf{Body-Part Phase Module}, which extracts structured, part-specific phase parameters (amplitude ($A$), frequency ($F$), phase shift ($S$), and offset ($B$)) from a reference motion. Second, a \textbf{Phase Encoder} maps this manifold into a compact, latent-aligned control embedding. Finally, rather than directly concatenating this code to the backbone, we introduce a \textbf{Phase ControlNet} that injects the phase signal through multi-layer residual modulation. This design structurally decouples the fine-grained control from the backbone motion modeling, enabling plug-and-play integration without altering the original generative architecture.

Crucially, this explicit parameterization naturally supports interactive, multi-round generation workflows. Users can extract the phase from a currently generated or reference motion, apply intuitive scalar edits to the target body parts, and re-apply the updated phase manifold to guide the next generation round until the desired behavior is achieved. Extensive experiments show that our method enables predictable, localized body-part control—such as adjusting movement magnitude, motion speed, or temporal timing—while largely preserving the remaining unedited parts of the motion. 
Benefiting from its modular design, the proposed method is highly effective across both diffusion- and flow-based backbones, demonstrating a plug-and-play capability to provide stable control without redesigning the underlying generator.
Overall, our results suggest that structured phase-based conditioning can serve as a highly practical control interface across diverse generative paradigms.

Our contributions can be summarized as follows:
\begin{itemize}
    \item We introduce a structured body-part phase representation that provides a compact and interpretable interface for localized motion control, marking the first application of phase parameters for body-part editing in generative models.
    \item We propose a modular Phase ControlNet framework that decouples structured control from backbone motion modeling, enabling plug-and-play integration via residual injection across different generative architectures (e.g., diffusion and flow).
    \item We conduct comprehensive experiments demonstrating that our method supports intuitive, interactive user edits, achieving stable and precise localized control without sacrificing the overall generation quality.
\end{itemize}

%% file: sec/2.related_work.tex
\section{Related Work}
\label{sec:related_work}

\subsection{Text-to-Motion Generation}
Text-to-motion (T2M) generation has made rapid progress in recent years, with many diffusion-based methods achieving strong motion quality and diversity \cite{mdm,t2mgpt,zhou2023let,jiang2023motiongpt,momask,fan2025go,xiao2025motionstreamer,barquero2024seamless,kong2023priority,zhou2024emdm,li2025genmo,zhang2024large,yuan2024mogents,bae2025less,pinyoanuntapong2025maskcontrol}. 
A common design is to first encode motions into a latent space using a motion VAE and then perform generation in the latent space, as exemplified by MLD~\cite{chen2023executing}. More recently, flow-based generative frameworks~\cite{wen2025hy-hunyuan,guo2025motionlab} have been explored as an alternative for motion generation with efficient sampling. 
Motivated by these advances, we propose a plug-and-play control module that can be integrated with both diffusion- and flow-based latent-space backbones to provide fine-grained, localized controllability.

\subsection{Controllable and Editable Text-to-Motion}

Beyond unconditional generation, a growing body of work investigates controllable T2M. 
Several approaches introduce explicit geometric constraints, such as joint trajectories or keyframes, into diffusion-based generators through condition injection, inpainting, or dedicated control branches~\cite{mdm,xie2023omnicontrol,dai2024motionlcm,karunratanakul2023guided,pinyoanuntapong2024mmm,wan2024tlcontrol,guo2025motionlab,pinyoanuntapong2025maskcontrol,liu2024programmable,bae2025less,hwang2025motion}. 
For example, OmniControl~\cite{xie2023omnicontrol} and MotionLCM~\cite{dai2024motionlcm} incorporate spatial--temporal control signals or trajectory encoders to guide selected joints during sampling. 
These methods provide strong geometric control, but the control interface is typically high-dimensional and requires dense, per-frame specifications, which may limit their usability for lightweight or interactive editing scenarios.
Another line of work focuses on text-driven motion editing~\cite{athanasiou2024motionfix,petrovich2024multi,li2025simmotionedit,guo2025motionlab}. 
Given a source motion and an edit prompt, these methods learn to generate a modified target motion that reflects the textual change. 
While offering an intuitive interaction modality, they often rely on large-scale paired datasets of source–target motions and corresponding edit descriptions. 
Moreover, the inherent ambiguity of natural language can make fine-grained, quantitatively predictable adjustments difficult to achieve.
In contrast to these approaches, our work seeks a structured and low-dimensional control space that enables precise, localized edits while remaining compatible with existing generative backbones.

\subsection{Phase-based Motion Representation}
Phase is a compact descriptor for periodic or quasi-periodic motion and has been widely used to parameterize motion progression. 
PFNN~\cite{holden2017phase} leverages a phase variable to modulate neural motion controllers for locomotion and motion in-between, enabling smooth interpolation across behaviors. 
Subsequent work such as DeepPhase~\cite{starke2022deepphase} learns multi-dimensional phase variables directly from unstructured motion capture data, providing a differentiable and interpretable representation of motion dynamics.
Phase has also been applied beyond motion control. WalkTheDog~\cite{li2024walkthedog} applies phase manifolds to cross-morphology motion alignment, demonstrating that phase representations capture intrinsic motion characteristics that transfer across different skeletal structures. 
However, most prior phase-based methods extract global, whole-body phase and are not designed for localized body-part control in text-to-motion generation. 
Our work builds on learned phase representations and introduces structured body-part phase as an explicit control interface for generative motion models.

%% file: sec/3.preliminaries.tex
\section{Preliminaries}
\label{sec:preliminaries}

\subsection{Diffusion-based Text-to-Motion in Latent Space}
We formulate diffusion-based text-to-motion generation in the motion latent space. 
Given a motion sequence $x$, a frozen motion VAE encoder $\mathcal{E}$ maps it to a latent code $z_0=\mathcal{E}(x)$. 
A forward noising process perturbs $z_0$ into $z_t$:
\begin{equation}
z_t = \sqrt{\bar{\alpha}_t}\, z_0 + \sqrt{1-\bar{\alpha}_t}\,\epsilon,\qquad 
\epsilon \sim \mathcal{N}(0,I),
\end{equation}
where $\{\bar{\alpha}_t\}$ is a pre-defined noise schedule and $t$ denotes the diffusion timestep.
We learn a conditional denoiser $\epsilon_\theta(z_t,t,c)$ with text embedding $c$ as condition, and train it with the standard noise-prediction objective:
\begin{equation}
\mathcal{L}_{\mathrm{DM}}(\theta)
=
\mathbb{E}_{t,\, z_0,\, \epsilon}
\left[\left\|\epsilon_\theta(z_t,t,c) - \epsilon\right\|_2^2\right].
\end{equation}
At inference time, we start from $z_T \sim \mathcal{N}(0,I)$ and iteratively denoise to obtain $\hat{z}_0$ using a diffusion sampler with $T$ steps (or a reduced number of steps with an accelerated sampler). 
Finally, the generated motion is decoded as $\hat{x}=\mathcal{D}(\hat{z}_0)$ via the frozen VAE decoder $\mathcal{D}$.

\subsection{Flow-based Text-to-Motion in Latent Space}
We follow a flow-matching formulation for text-to-motion generation in the latent space of a pretrained motion VAE. 
Given a motion sequence $x$, a frozen encoder $\mathcal{E}$ maps it to $z_0=\mathcal{E}(x)$, while a source latent is sampled as $z_1\sim\mathcal{N}(0,I)$. 
We define a linear coupling path
\begin{equation}
z_t=(1-t)z_0+t z_1,\qquad t\in[0,1],
\end{equation}
and learn a text-conditioned velocity field $v_\theta(z_t,t,c)$ (with text embedding $c$) by matching the target velocity along this path:
\begin{equation}
\mathcal{L}_{\mathrm{FM}}(\theta)
=
\mathbb{E}_{t\sim\mathcal{U}[0,1],\, z_0,\, z_1}
\left[\left\|v_\theta(z_t,t,c) - (z_1-z_0)\right\|_2^2\right].
\end{equation}
At inference, we initialize $z_1\sim\mathcal{N}(0,I)$ and integrate the ODE $\frac{dz}{dt}=v_\theta(z,t,c)$ backward from $t=1$ to $t=0$ using $N$ steps (e.g., Euler), yielding $\hat{z}_0$, which is decoded as $\hat{x}=\mathcal{D}(\hat{z}_0)$ with the frozen decoder $\mathcal{D}$. 
This latent flow backbone also provides a natural place to attach additional conditioning interfaces; in the following, we introduce a compact body-part phase interface for structured, localized motion control.

%% file: sec/4_method_fk.tex
\section{Method}
\label{sec:method}

\input{fig_show/show_pipeline}

\subsection{Overview}
We propose \textbf{Modular Body-Part Phase Control}, a plug-and-play framework for controllable text-to-motion generation. As illustrated in Fig.~\ref{fig:pipeline}, our approach augments a pretrained latent-space motion generator with an external phase control module, deliberately avoiding any modifications to the original backbone architecture. The framework comprises three core components: (i) a \textbf{Body-Part Phase Module} that extracts localized, periodic phase parameters from a reference motion to establish a time-dependent phase manifold; (ii) a \textbf{Phase Encoder} that projects this structured manifold into a latent-aligned conditioning space; and (iii) a \textbf{Phase ControlNet} that injects multi-layer modulation residuals into the backbone to guide the generation process. 

By structurally decoupling the fine-grained phase control from the backbone motion modeling, our method enables localized body-part edits via a unified interface. Furthermore, because our control module interacts with the backbone solely through additive residual injection, it can be seamlessly integrated with both diffusion and flow-matching backbones (as formulated in Sec.~\ref{sec:preliminaries}).

\subsection{Body-Part Phase Module and User Controls}
\label{sec:phase_representation_controls}
Human motion inherently exhibits quasi-periodic patterns within local temporal windows, often characterized by distinct rhythms and timings across different body parts. To systematically capture this localized periodic structure, we partition the human skeleton into a semantic set of parts:
We split the human body into a small set of high-level parts
$\mathcal{B}=\{\texttt{left\_up}$,$\texttt{right\_up}$,\texttt{left\_low}, $\texttt{right\_low}$, $\texttt{trunk}\}$.

\textbf{Body-Part Phase Module.}
For a given part $b\in\mathcal{B}$, we assume its 1D kinematic signal $v_b(\tau)$ (e.g., a joint velocity trajectory) over a normalized temporal window $\tau\in[0,1]$ can be compactly approximated by a set of $K$ sinusoidal basis functions~\cite{starke2022deepphase}:
\begin{equation}
v_b(\tau) \approx \sum_{k=1}^{K} A_{b,k}\cos\!\big(2\pi(F_{b,k}\tau + S_{b,k})\big) + B_{b,k},
\end{equation}
where $A_{b,k}$, $F_{b,k}$, $S_{b,k}$, and $B_{b,k}$ correspond to the amplitude, frequency, phase shift, and static offset, respectively. These parameters offer highly disentangled control semantics: $A$ governs the motion intensity, $F$ dictates the execution pace or repetition rate, and $S$ adjusts the temporal alignment. We utilize a \textbf{Body-Part Phase Extractor}, built upon a periodic autoencoder architecture~\cite{starke2022deepphase}, to accurately estimate these parameters.

Subsequently, we map these scalar parameters into a continuous, time-dependent \textbf{phase manifold} via sinusoidal embeddings:
\begin{equation}
\phi_{b,k}(\tau)=2\pi(\tau\cdot F_{b,k}+S_{b,k}),\qquad
m_{b,k}(\tau)=A_{b,k}\,[\cos\phi_{b,k}(\tau),\,\sin\phi_{b,k}(\tau)].
\end{equation}
Concatenating the embeddings $m_{b,k}$ across all $(b,k)$ pairs yields a comprehensive phase feature per time step. Stacking these features over the target temporal window forms the phase sequence $\mathbf{P}$, which serves as the foundational conditioning signal for the downstream generative process.

\textbf{Interactive User Controls.}
The explicit parameterization of the phase manifold enables intuitive and localized motion editing. Given a target body part $b^\star$, users can apply simple scalar manipulations to its corresponding parameters: scaling amplitude $A_{b^\star,k}$ to adjust movement magnitude, scaling frequency $F_{b^\star,k}$ to modify the motion pace, or shifting $S_{b^\star,k}$ to alter event timing. Following user edits, the updated parameters are seamlessly re-embedded into a modified phase sequence $\mathbf{P}$, facilitating multi-round, interactive motion refinement.

\subsection{Phase Encoder and ControlNet Injection}
\label{sec:phase_encoder_controlnet}
To enforce the desired spatial-temporal control without disrupting the prior of the pretrained motion backbone, we introduce an auxiliary conditioning branch. This branch projects the structured phase sequence $\mathbf{P}\in\mathbb{R}^{T\times 20}$ into the backbone's feature space via a \textbf{Phase Encoder} and performs multi-layer residual injection using a \textbf{Phase ControlNet}.

\textbf{Phase Encoder.}
We employ a lightweight, trainable temporal encoder $\mathcal{E}_{\psi}$ to map the phase sequence into a compact, latent-aligned control embedding:
$
g = \mathcal{E}_{\psi}(\mathbf{P}).
$
Architecturally, $\mathcal{E}_{\psi}$ comprises a small stack of 1D temporal convolutions followed by a projection head. This design effectively bridges the continuous phase space with the latent representation domain of the motion backbone.

\textbf{Phase ControlNet.}
Let $\mathcal{G}_\theta$ denote the backbone generator (Sec.~\ref{sec:preliminaries}) composed of $L$ blocks, which processes the input tuple $(z_t, t, c)$ to predict the denoised latent, exposing intermediate feature maps $\{h^{(\ell)}\}_{\ell=1}^{L}$. We instantiate the Phase ControlNet $\mathcal{C}_{\eta}$ to exactly mirror the block architecture of $\mathcal{G}_\theta$.

The control embedding $g$ is first integrated into the generation pipeline by projecting it via a multi-layer perceptron (MLP) and adding it to the current noised latent:
$
\tilde{z}_t = z_t + \mathrm{MLP}(g).
$
The Phase ControlNet then processes this augmented latent to generate a set of block-wise residuals:
\begin{equation}
\{r^{(\ell)}\}_{\ell=1}^{L} = \mathcal{C}_{\eta}(\tilde{z}_t,t,c),
\end{equation}
where $L=9$ in our implementation. Crucially, each residual $r^{(\ell)}$ is computed via a zero-initialized projection layer from the corresponding ControlNet block. This initialization ensures that $\mathcal{C}_{\eta}$ initially exerts zero influence, preserving the generation quality of the backbone at the onset of training. During the forward pass, these residuals are additively injected into the backbone blocks:
\begin{equation}
h^{(\ell)} \leftarrow h^{(\ell)} + r^{(\ell)}, \qquad \ell=1,\dots,L.
\end{equation}
This layer-wise modulation enables stable and localized guidance of specific body parts while largely preserving the synthesized dynamics of the remaining unedited parts.

\subsection{Training and Inference}
\label{sec:training_inference}
We adopt a modular, two-stage training paradigm. Throughout both stages, the core generative prior—comprising the motion VAE, text encoder, and latent-space backbone generator—is strictly frozen.

\textbf{Stage I: Body-Part Phase Module Pretraining.}
We first train the phase network $\{\Phi_b\}_{b\in\mathcal{B}}$ independently to estimate and reconstruct periodic parameters from joint-velocity trajectories $v_b$ within short windows. This is optimized via a standard reconstruction loss:
\begin{equation}
\mathcal{L}_{\mathrm{phase\_pre}} = \sum_{b\in\mathcal{B}} \left\|\hat{v}_b - v_b\right\|_2^2.
\end{equation}
Once trained, the phase networks $\{\Phi_b\}$ are kept frozen for subsequent operations.

\textbf{Stage II: Phase Encoder and ControlNet Training.}
In the second stage, we jointly optimize the Phase Encoder $\mathcal{E}_{\psi}$ and the Phase ControlNet $\mathcal{C}_{\eta}$ using a composite objective:
\begin{equation}
\mathcal{L} = \mathcal{L}_{\mathrm{DM}} + \lambda_{\mathrm{phase}}\mathcal{L}_{\mathrm{phase}},
\end{equation}
where $\mathcal{L}_{\mathrm{DM}}$ is the standard generative objective (e.g., noise-prediction loss). To enforce high-fidelity phase alignment, we introduce an auxiliary phase consistency loss $\mathcal{L}_{\mathrm{phase}}$. We decode the predicted denoised latent $\hat{z}_0$ into motion features $\hat{x}$, extract its phase sequence via the frozen $\{\Phi_b\}$, and compute a masked $L_1$ distance against the phase sequence of the reference motion $x_{\mathrm{ref}}$:
\begin{equation}
\mathcal{L}_{\mathrm{phase}} = \frac{1}{\sum \mathbf{M}} \sum \mathbf{M}\odot \left\|\mathbf{P}(\hat{x})-\mathbf{P}(x_{\mathrm{ref}})\right\|_{1},
\end{equation}
where $\mathbf{P}(x)=\Psi\!\left(\{\Phi_b(x)\}_{b\in\mathcal{B}}\right)$, $\Psi(\cdot)$ denotes the sinusoidal phase manifold construction (Sec.~\ref{sec:phase_representation_controls}), and $\mathbf{M}$ is a spatio-temporal mask focusing the penalty on the specific edited parts.

\textbf{Inference Pipeline.}
During inference, the unified framework delivers highly controllable generation. Given a reference motion $x_{\mathrm{ref}}$ and a text prompt, we first extract the base text embedding $c$ and the base phase parameters using $\{\Phi_b\}$. Users seamlessly edit the target body parts via scalar modifications to $(A, F, S)$, from which the updated phase sequence $\mathbf{P}$ is reconstructed. This sequence is mapped to the control embedding $g = \mathcal{E}_{\psi}(\mathbf{P})$ and passed to the Phase ControlNet $\mathcal{C}_{\eta}$. As the backbone generator iteratively denoises the latent $z_t$ conditioned on $c$, the ControlNet continuously injects the structural residuals $r^{(\ell)}$. The final denoised latent $\hat{z}_0$ is then passed through the motion VAE decoder to yield the synthesized motion $\hat{x} = \mathcal{D}(\hat{z}_0)$, which strictly adheres to the textual description while accurately reflecting the localized, user-defined rhythmic adjustments.

%% file: fig_show/show_pipeline.tex
\begin{figure}[t]
    \centering
    \includegraphics[width=0.7\linewidth]{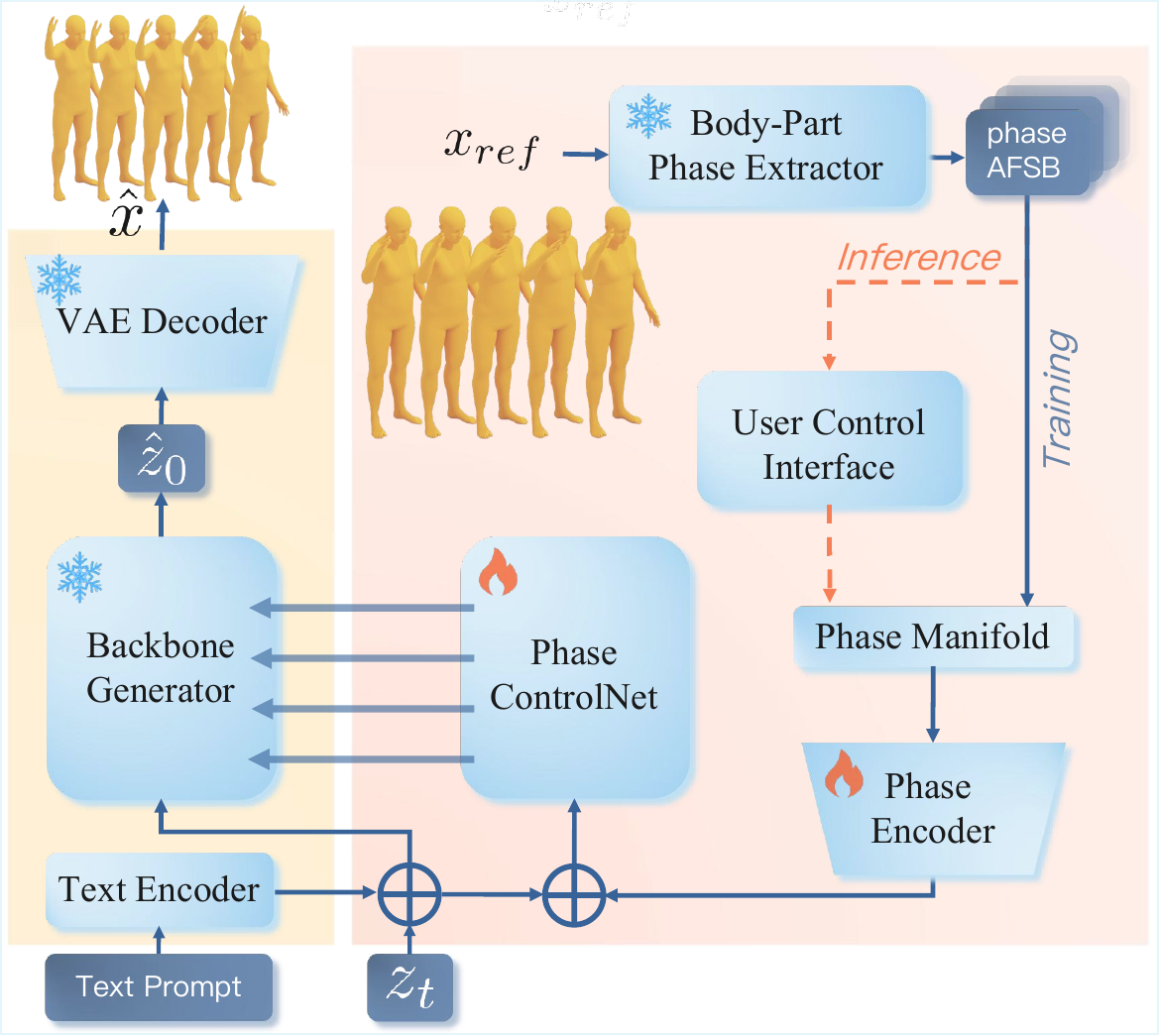}
    \caption{Overview of our modular body-part phase control framework. Given a reference motion, a frozen body-part phase extractor predicts per-part periodic parameters (AFSB) of each body part.
    A Phase ControlNet then injects multi-layer residuals into the backbone generator to produce motion latents aligned with the periodic parameters.
    The generated latent is finally decoded by a frozen motion VAE decoder to obtain the final motion.
    Users can interactively edit these parameters via simple scalar controls, which are converted to a phase manifold and encoded into a control embedding.
    }
    \label{fig:pipeline}
\end{figure}

%% file: sec/5.experiments.tex
\section{Experiments}
\label{sec:experiments}

\subsection{Experimental Setup}
\label{sec:exp_setup}
\textbf{Datasets.}
We experiment on the popular HumanML3D~\cite{guo2022generating-HumanML3D} dataset, featuring 14,616 unique human motion sequences with 44,970 textual descriptions. 
For a fair comparison with previous methods~\cite{chen2023executing,petrovich2022temos,mdm,zhang2024motiondiffuse,guo2022generating-HumanML3D,dai2024motionlcm}, we take the redundant motion representation, including root velocity, root height, local joint positions, velocities, rotations in root space, and the foot-contact binary labels.

\noindent\textbf{Evaluation metrics.}
We extend the evaluation metrics of previous works~\cite{chen2023executing,guo2022generating-HumanML3D,xie2023omnicontrol,dai2024motionlcm}.
(1) \textbf{Time cost:} We follow~\cite{chen2023executing} to report the \underline{A}verage \underline{I}nference \underline{T}ime per \underline{S}entence (AITS) to evaluate the inference efficiency of models.
(2) \textbf{Motion quality:} Fr\'echet Inception Distance (FID) is adopted as a principal metric to evaluate the feature distributions between the generated and real motions. The feature extractor employed is from~\cite{guo2022generating-HumanML3D}.
(3) \textbf{Motion diversity:} MultiModality (MModality) measures the generation diversity conditioned on the same text and Diversity calculates variance through features~\cite{guo2022generating-HumanML3D}.
(4) \textbf{Condition matching:} Following~\cite{guo2022generating-HumanML3D}, we calculate the motion-retrieval precision (R-Precision) to report the text-motion Top-1/2/3 matching accuracy and Multimodal Distance (MM Dist) calculates the mean distance between motions and texts.

\noindent\textbf{Implementation details.} 
We utilize MotionLCM~\cite{dai2024motionlcm} (one-step sampling, CFG $w{=}7.5$) for our diffusion backbone, and a latent rectified-flow network trained via flow-matching for our flow backbone and train a velocity-field network with the flow-matching objective. 
Sampling is performed by ODE integration with a fixed discretization of $N$ steps.
We train the Phase Encoder and Phase ControlNet using AdamW, cosine learning-rate decay, and 1K-step linear warm-up, with batch size 128 and learning rate $1\times10^{-4}$.
The Phase ControlNet is initialized from the corresponding backbone weights and uses zero-initialized projections to output $L{=}9$ block-aligned residuals, which are additively injected after each backbone block.
We use $K{=}2$ phase channels for each of $|\mathcal{B}|{=}5$ body parts.
All models are implemented in PyTorch and trained on an NVIDIA RTX 4090 GPU, with evaluation on a Tesla V100 GPU.

\subsection{Quantitative Results}

\input{excel_show/main_quant_compare}
\input{fig_show/show_AF}
\input{excel_show/change_AFS_excel}

We first evaluate our method on the text-to-motion (T2M) task, with results summarized in Table~\ref{tab:main_quant_compare}. 
We show our method's results on top of both diffusion and flow backbones.
Besides, we compare our method against representative T2M baselines on HumanML3D~\cite{guo2022generating-HumanML3D} using the standard evaluation protocol and metrics recommended in prior work~\cite{guo2022generating-HumanML3D}. All metrics are reported with a 95\% confidence interval computed over 20 independent runs. Following MotionLCM, we use the same evaluation setup, and baseline numbers are taken from the MotionLCM report for a fair comparison. Overall, our method achieves consistently strong performance across multiple metrics, including R-Precision, FID, MM Dist, and Diversity. 
We attribute these gains to the additional phase cues provided by the phase-conditioned ControlNet, which supply informative motion priors beyond text, thereby improving alignment and motion realism.

\noindent\textbf{Runtime Efficiency.} We also analyze the runtime efficiency of our method. As shown in Table~\ref{tab:main_quant_compare}, when integrated into MotionLCM, our phase-conditioned ControlNet increases AITS by only 0.004\,s compared to MotionLCM. This marginal latency stems from the lightweight design of our Body-Part Phase Extractor and Phase Encoder modules.

\noindent\textbf{Control-Response Correlation Analysis.} 
We quantitatively assess the correlation between input controls and motion responses using the amplitude/frequency scale response curves in Fig.~\ref{fig:phase_stat}. 
The x-axis denotes the user-specified scale factor $x$, and the y-axis reports the effective ratio $(X'/X)$ between the generated and reference motions. 
For each motion case, we average $(X'/X)$ over multiple samples, and then compute the mean and standard deviation across all cases. 
Within the typical editing range ($0.5 \le x \le 1.5$), $(X'/X)$ follows $x$ with an almost linear trend and low variance, indicating proportional and precise localized control.


\noindent\textbf{Motion Quality after Edition.} 
As shown in Table~\ref{tab:change_AFS}, we conduct controllable generation experiments by manipulating three phase-related attributes: amplitude (A), frequency (F), and phase shift (S). 
All variants are built upon MotionLCM~\cite{dai2024motionlcm} with our phase-conditioned ControlNet. We set the control strength to 1.5 for amplitude and frequency, and 0.5 for phase shift. 
For each attribute (A/F/S), we report results for global control as well as two representative body-part settings: left lower limb (LL) and right upper limb (RU), leveraging left--right symmetry. 
Full results for all body parts are provided in the supplementary material. 
The results indicate that, under controllable editing, our method maintains quantitative scores close to the unedited setting, demonstrating that it enables effective control while preserving overall motion quality.

\input{fig_show/show_change_AFS}

\subsection{Qualitative Results}
To visually demonstrate the fine-grained, localized editing capabilities of our Modular Body-Part Phase Control, we present qualitative examples of modifying distinct phase attributes (Phase Shift, Amplitude, and Frequency) on specific body parts, as illustrated in Fig.~\ref{fig:cahnge_AFS}. 
For all comparisons, the middle row (d, e, f) displays the original generated motions conditioned solely on the text prompts, while the top and bottom rows show the results with modified phase parameters.

\noindent\textbf{Phase Shift (Timing) Control.} 
The first column (a, d, g) shows motions generated from the prompt \textit{``the character scratches his head with his right arm.''}
We apply phase shift ($S$) edits exclusively to the right upper limb. 
By decreasing the phase shift ($S - 0.125$) in (a), the scratching action is initiated earlier in the temporal sequence. 
Conversely, increasing the phase shift ($S + 0.125$) in (g) delays the execution of the gesture. 
Throughout this temporal shifting, the global posture and the semantics of the scratching action remain structurally intact and natural.

\noindent\textbf{Amplitude (Magnitude) Control.} 
The second column (b, e, h) illustrates amplitude modification for the prompt \textit{``a person is waving with his right hand.''}.
By scaling the amplitude ($A$) of the right upper limb, we directly control the spatial magnitude of the waving gesture. 
In (b), scaling by $A \times 0.5$ results in a subtle, restrained hand wave. 
In contrast, scaling by $A \times 1.5$ in (h) produces a highly exaggerated, wide-ranging arm swing. 
Notably, the unedited body parts (e.g., the lower limbs and trunk) remain strictly coherent with the baseline (e), demonstrating the excellent disentanglement of our part-level control.

\noindent\textbf{Frequency (Pace) Control.} 
In the third column (c, f, i), we demonstrate frequency control over cyclic locomotion using the prompt \textit{``person carefully walks with left right first in a straight direction.''} Here, we target both lower limbs. Scaling the frequency by $F \times 0.5$ in (c) explicitly slows down the stepping pace, resulting in a cautious, slow-motion stride. 
Alternatively, doubling the frequency ($F \times 2.0$) in (i) significantly accelerates the leg repetition rate, turning the walk into a brisk, rapid stepping motion. 
The structural integrity of the upper body and the semantic ``straight direction'' are perfectly preserved.


Overall, these qualitative results corroborate our quantitative findings. 
They confirm that our phase-based interface enables predictable, independent, and interpretable control over specific body parts without compromising the global quality of the generative prior.

\subsection{Ablations}
\input{excel_show/Ablation_excel}

We conduct ablation experiments on HumanML3D to analyze (i) the effectiveness of introducing phase signals via our phase-conditioned ControlNet design, and (ii) the impact of phase conditioning granularity.

\noindent\textbf{Effectiveness of phase-conditioned ControlNet.}
We first examine a straightforward conditioning strategy that directly concatenates body-part phase features to the flow network inputs, and compare it with our flow-based phase-conditioned ControlNet. 
Visualization results indicate that changing the phase condition does not reliably induce corresponding changes in the generated motion, suggesting that the flow model tends to ignore the phase signal when it is provided only as an auxiliary input.
As shown in Table~\ref{tab:Ablation}, our method achieves consistently better performance on key metrics such as R-Precision and MM Dist, while maintaining competitive FID and Diversity. 
These gains mainly stem from ControlNet-style conditioning, which injects phase cues through dedicated modulation pathways and prevents them being ignored as weak auxiliary inputs.

\noindent\textbf{Impact of phase granularity.}
We further investigate the impact of phase granularity by comparing a whole-body (non-partitioned) phase representation with our body-part phase decomposition, where both variants are implemented on MotionLCM using the same phase-conditioned ControlNet. 
Our body-part phase decomposition consistently outperforms the whole-body variant across all metrics, as it provides finer-grained part-wise dynamics and avoids averaging out localized dynamics.

%% file: excel_show/main_quant_compare.tex

\begin{table*}[t]
\centering
\caption{Comparison of text-conditional motion synthesis on the HumanML3D dataset. We compute evaluation metrics following the standard protocol and report the average over multiple runs with a 95\% confidence interval. ``$\rightarrow$'' indicates that closer to real data is better. \textbf{Bold} and \underline{underline} indicate the best and second best results.}
\label{tab:main_quant_compare}
\resizebox{\textwidth}{!}{%
\begin{tabular}{lccccccc}
\toprule
\multirow{2}{*}{\textbf{Methods}} & \multirow{2}{*}{\textbf{AITS}$\downarrow$} & \multicolumn{3}{c}{\textbf{R-Precision}$\uparrow$} & \multirow{2}{*}{\textbf{FID}$\downarrow$} & \multirow{2}{*}{\textbf{MM Dist}$\downarrow$} & \multirow{2}{*}{\textbf{Diversity}$\rightarrow$} \\
\cmidrule(lr){3-5}
 &  & \textbf{Top 1} & \textbf{Top 2} & \textbf{Top 3} &  &  &  \\
\midrule
Real & -- & $0.511\pms{.003}$ & $0.703\pms{.003}$ & $0.797\pms{.002}$ & $0.002\pms{.000}$ & $2.974\pms{.008}$ & $9.503\pms{.065}$ \\
\midrule
Seq2Seq~\cite{lin2018generating} & -- & $0.180\pms{.002}$ & $0.300\pms{.002}$ & $0.396\pms{.002}$ & $11.75\pms{.035}$ & $5.529\pms{.007}$ & $6.223\pms{.061}$ \\
JL2P~\cite{ahuja2019language2pose} & -- & $0.246\pms{.002}$ & $0.387\pms{.002}$ & $0.486\pms{.002}$ & $11.02\pms{.046}$ & $5.296\pms{.008}$ & $7.676\pms{.058}$ \\
T2G~\cite{bhattacharya2021text2gestures} & -- & $0.165\pms{.001}$ & $0.267\pms{.002}$ & $0.345\pms{.002}$ & $7.664\pms{.030}$ & $6.030\pms{.008}$ & $6.409\pms{.071}$ \\
Hier~\cite{ghosh2021synthesis} & -- & $0.301\pms{.002}$ & $0.425\pms{.002}$ & $0.552\pms{.004}$ & $6.532\pms{.024}$ & $5.012\pms{.018}$ & $8.332\pms{.042}$ \\
TEMOS~\cite{petrovich2022temos} & $0.017$ & $0.424\pms{.002}$ & $0.612\pms{.002}$ & $0.722\pms{.002}$ & $3.734\pms{.028}$ & $3.703\pms{.008}$ & $8.973\pms{.071}$ \\
T2M~\cite{guo2022generating-HumanML3D} & $0.038$ & $0.457\pms{.002}$ & $0.639\pms{.003}$ & $0.740\pms{.003}$ & $1.067\pms{.002}$ & $3.340\pms{.008}$ & $9.188\pms{.002}$ \\
MDM~\cite{mdm} & $24.74$ & $0.320\pms{.005}$ & $0.498\pms{.004}$ & $0.611\pms{.007}$ & $0.544\pms{.044}$ & $5.566\pms{.027}$ & $\underline{9.559}\pms{.086}$ \\
MotionDiffuse~\cite{zhang2024motiondiffuse} & $14.74$ & $0.491\pms{.001}$ & $0.681\pms{.001}$ & $0.782\pms{.001}$ & $0.630\pms{.001}$ & $3.113\pms{.001}$ & $9.410\pms{.049}$ \\
MLD~\cite{chen2023executing} & $0.217$ & $0.481\pms{.003}$ & $0.673\pms{.003}$ & $0.772\pms{.002}$ & $0.473\pms{.013}$ & $3.196\pms{.010}$ & $9.724\pms{.082}$ \\
MotionLCM~\cite{dai2024motionlcm} & $\textbf{0.030}$ & $0.502\pms{.003}$ & $\underline{0.701}\pms{.002}$ & $\underline{0.803}\pms{.002}$ & $0.467\pms{.012}$ & $3.022\pms{.009}$ & $9.631\pms{.066}$ \\


\midrule
Ours~(Flow) &0.157 &$\mathbf{0.531}\pms{.006}$ &$\mathbf{0.719}\pms{.004}$ &$\mathbf{0.808}\pms{.005}$ &$\mathbf{0.146}\pms{.007}$ &$\mathbf{2.880}\pms{.018}$ &$9.318\pms{.055}$\\

Ours~(MotionLCM) & \underline{0.034} & $\underline{0.503}\pms{.005}$ & $\underline{0.701}\pms{.005}$ & $0.796\pms{.004}$ &$\underline{0.465}\pms{.022}$ &$\underline{3.000}\pms{.013}$ &$\mathbf{9.496}\pms{.078}$\\

\bottomrule
\end{tabular}%
}
\end{table*}

%% file: fig_show/show_AF.tex


%

\begin{figure}[t]
  \centering
  \includegraphics[width=0.8\textwidth]{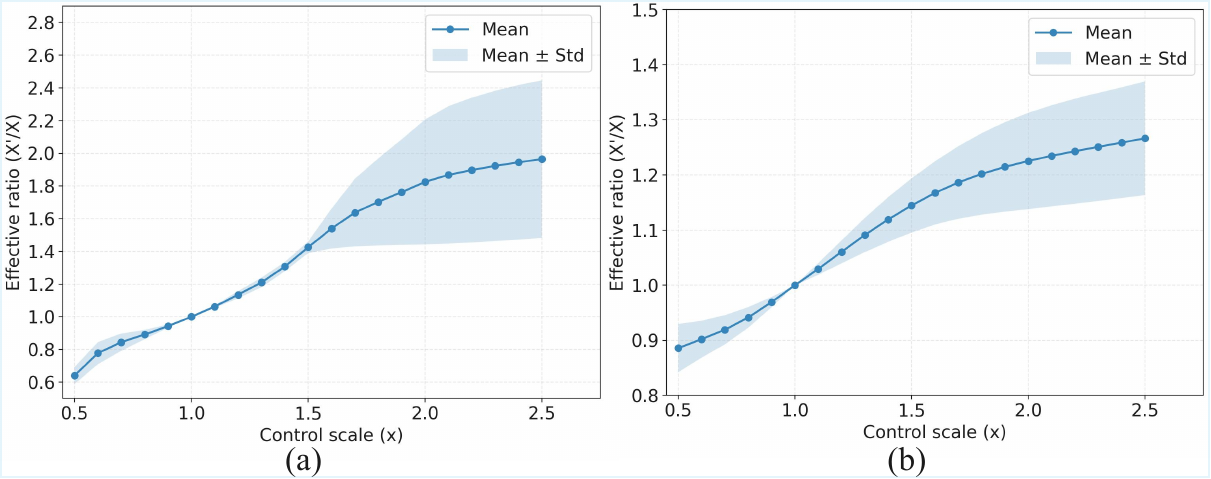}
  \caption{
  \textbf{Control-response correlation curves.} 
  Scale response curves for (a) amplitude and (b) frequency control. 
  The x-axis represents the explicit control scale factor applied by the user, and the y-axis denotes the measured effective ratio ($X'/X$) of the generated motion relative to the reference. 
  The solid blue line represents the global mean computed via hierarchical aggregation across all test cases, while the shaded region indicates the standard deviation. 
  The curves show a highly proportional linear correlation within the typical editing range ($0.5 \le x \le 1.5$). 
  At extreme scales, they transition into a sub-linear regime with higher variance, reflecting the physical plausibility constraints of the generative prior.
  }
  \label{fig:phase_stat}
\end{figure}

%% file: excel_show/change_AFS_excel.tex
\begin{table*}[t]
\centering
\caption{Quantitative results of amplitude (A), frequency (F) and phase shift (S) control on HumanML3D. All variants are built upon MotionLCM with our phase-conditioned ControlNet. The control strength is set to $1.5$ for A and F, and $0.5$ for S.  For each attribute (A/F/S), we report results on global control and the left lower limb (LL) and right upper limb (RU) due to left-right symmetry. Full results are provided in the supplementary material.}
\label{tab:change_AFS}
\resizebox{0.9\textwidth}{!}{%
\begin{tabular}{lcccccc}
\toprule
\multirow{2}{*}{\textbf{Methods}} & \multicolumn{3}{c}{\textbf{R-Precision}$\uparrow$} & \multirow{2}{*}{\textbf{FID}$\downarrow$} & \multirow{2}{*}{\textbf{MM Dist}$\downarrow$} & \multirow{2}{*}{\textbf{Diversity}$\rightarrow$} \\
\cmidrule(lr){2-4}
 & \textbf{Top 1} & \textbf{Top 2} & \textbf{Top 3} &  &  &  \\
\midrule
Real & $0.511\pms{.003}$ & $0.703\pms{.003}$ & $0.797\pms{.002}$ & $0.002\pms{.000}$ & $2.974\pms{.008}$ & $9.503\pms{.065}$ \\

\midrule

Base (no scaling) & $0.503\pms{.005}$ & $0.701\pms{.005}$ & $0.796\pms{.004}$ & $0.465\pms{.022}$ & $3.000\pms{.013}$ & $9.496\pms{.078}$\\

\midrule

A-global & $0.507\pms{.006}$ & $0.705\pms{.004}$ & $0.799\pms{.005}$ & $0.305\pms{.014}$ & $2.971\pms{.018}$ & $9.489\pms{.077}$\\
A-LL & $0.504\pms{.005}$ & $0.699\pms{.006}$ & $0.796\pms{.005}$ & $0.428\pms{.022}$ & $3.005\pms{.015}$ & $9.514\pms{.079}$\\
A-RU & $0.503\pms{.006}$ & $0.701\pms{.005}$ & $0.795\pms{.005}$ & $0.390\pms{.019}$ & $2.982\pms{.016}$ & $9.470\pms{.082}$\\

\midrule

F-global & $0.481\pms{.007}$ & $0.669\pms{.006}$ & $0.770\pms{.005}$ & $0.340\pms{.014}$ & $3.144\pms{.021}$ & $9.393\pms{.076}$\\
F-LL & $0.501\pms{.005}$ & $0.699\pms{.006}$ & $0.795\pms{.006}$ & $0.345\pms{.021}$ & $3.004\pms{.016}$ & $9.408\pms{.071}$\\
F-RU & $0.502\pms{.006}$ & $0.698\pms{.005}$ & $0.794\pms{.004}$ & $0.354\pms{.018}$ & $3.003\pms{.016}$ & $9.492\pms{.083}$\\

\midrule

S-global & $0.484\pms{.005}$ & $0.675\pms{.006}$ & $0.775\pms{.006}$ & $0.427\pms{.016}$ & $3.120\pms{.015}$ & $9.297\pms{.082}$\\
S-LL & $0.462\pms{.006}$ & $0.652\pms{.006}$ & $0.753\pms{.005}$ & $0.843\pms{.028}$ & $3.343\pms{.023}$ & $9.049\pms{.081}$\\
S-RU & $0.485\pms{.006}$ & $0.677\pms{.005}$ & $0.778\pms{.006}$ & $0.486\pms{.022}$ & $3.100\pms{.018}$ & $9.263\pms{.088}$\\

\bottomrule
\end{tabular}%
}
\end{table*}

%% file: fig_show/show_change_AFS.tex
\begin{figure}[t]
    \centering
    \includegraphics[width=0.8\linewidth]{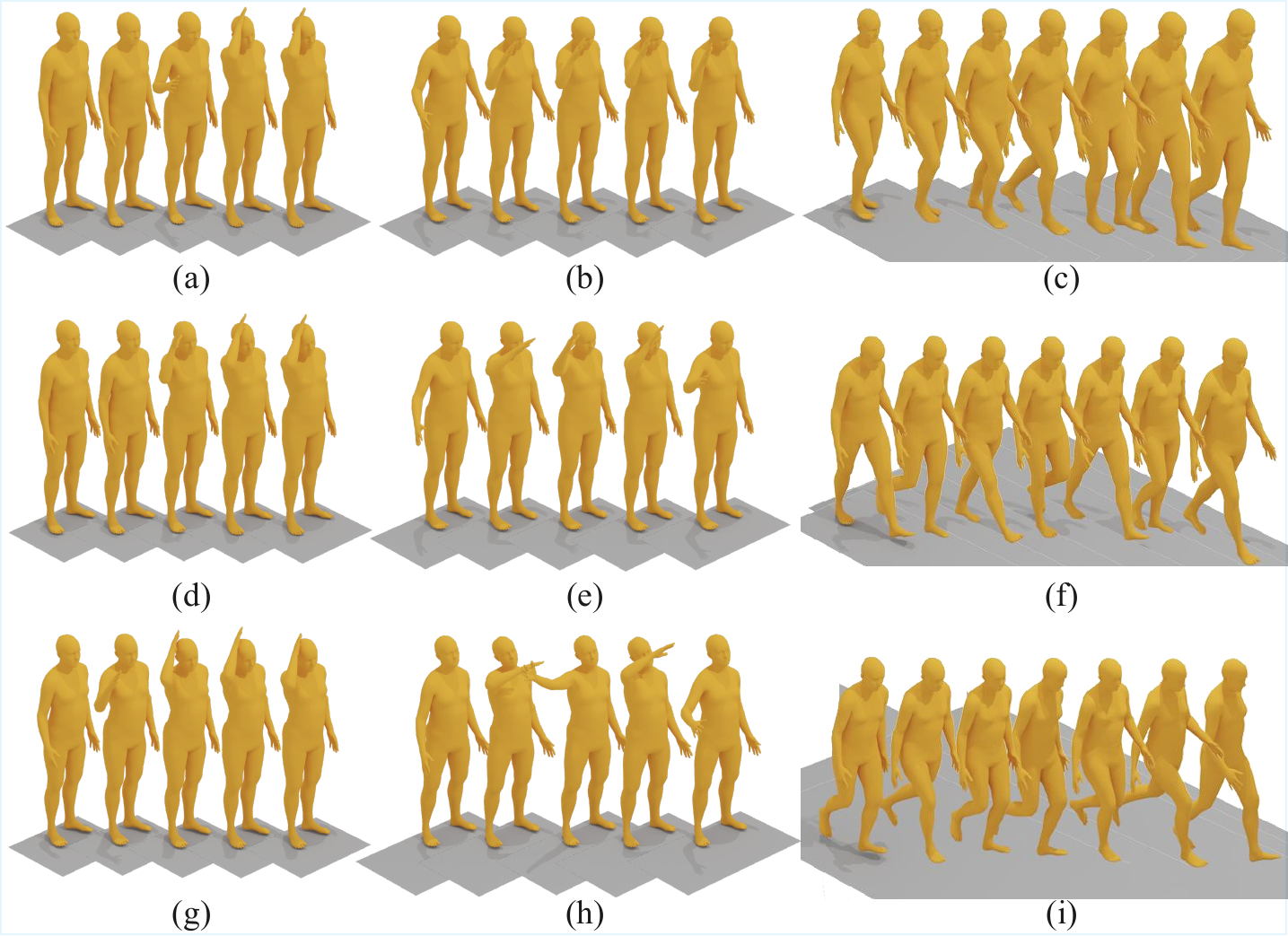}
    \caption{
    \textbf{Qualitative results of body-part motion editing.} 
    The middle row (d, e, f) displays the original motions generated from text prompts. 
    The top and bottom rows show the results after applying our modular phase control. 
    \textbf{(a, d, g):} Adjusting the shift ($S$) of the right arm to alter the timing of a scratching gesture. 
    \textbf{(b, e, h):} Scaling the amplitude ($A$) of the right arm to control the magnitude of a waving action. 
    \textbf{(c, f, i):} Scaling the frequency ($F$) of both legs to change the stepping pace of a walk. 
    Our method precisely edits the target parts while keeping the rest of the body coherent.
    }
    \label{fig:cahnge_AFS}
\end{figure}

%% file: excel_show/Ablation_excel.tex
\begin{table*}[t]
\centering
\caption{Ablation study on HumanML3D. We compare naive phase conditioning and full-body (non-partitioned) phase control with our body-part phase control.}
\label{tab:Ablation}
\resizebox{0.9\textwidth}{!}{%
\begin{tabular}{l ccc ccc}
\toprule
\multirow{2}{*}{\textbf{Methods}} & \multicolumn{3}{c}{\textbf{R-Precision}$\uparrow$} & \multirow{2}{*}{\textbf{FID}$\downarrow$} & \multirow{2}{*}{\textbf{MM Dist}$\downarrow$} & \multirow{2}{*}{\textbf{Diversity}$\rightarrow$} \\
\cmidrule(lr){2-4}
& \textbf{Top 1} & \textbf{Top 2} & \textbf{Top 3} &  &  & \\
\midrule
Real & $0.511\pms{.003}$ & $0.703\pms{.003}$ & $0.797\pms{.002}$ & $0.002\pms{.000}$ & $2.974\pms{.008}$ & $9.503\pms{.065}$ \\

\midrule
Flow~(Phase Cond.) & $0.524\pms{.003}$ &$0.718\pms{.003}$ &$\mathbf{0.815}\pms{.002}$ &$\mathbf{0.100}\pms{.003}$ &$2.886\pms{.008}$ &$\mathbf{9.337}\pms{.089}$\\

Ours~(Flow) &$\mathbf{0.531}\pms{.006}$ &$\mathbf{0.719}\pms{.004}$ &$0.808\pms{.005}$ &$0.146\pms{.007}$ &$\mathbf{2.880}\pms{.018}$ &$9.318\pms{.055}$\\

\midrule

Ours~(MotionLCM, Whole) & $0.467\pms{.006}$ & $0.653\pms{.005}$ & $0.752\pms{.005}$ & $1.287\pms{.033}$ & $3.310\pms{.021}$ & $9.103\pms{.084}$\\

Ours~(MotionLCM) & $\mathbf{0.503}\pms{.005}$ & $\mathbf{0.701}\pms{.005}$ & $\mathbf{0.796}\pms{.004}$ &$\mathbf{0.465}\pms{.022}$ &$\mathbf{3.000}\pms{.013}$ &$\mathbf{9.496}\pms{.078}$\\

\bottomrule
\end{tabular}%
}
\end{table*}

%% file: sec/6.conclusion.tex
\section{Conclusion and Limitations}
\label{sec:conclusion}

We present \textbf{Modular Body-Part Phase Control}, a plug-and-play framework for controllable text-to-motion generation. 
Our approach extracts a structured body-part phase signal from a reference motion and injects it into a latent-space generator through a Phase ControlNet with multi-layer residual modulation. This design provides an interpretable and user-friendly control interface based on simple scalar inputs, enabling predictable and localized body-part adjustments in motion magnitude, speed or repetition rate, and timing, while largely preserving overall motion coherence. Experiments on diffusion- and flow-based backbones demonstrate that our module improves controllability without sacrificing generation quality and supports interactive multi-round refinement.

\textbf{Limitations.} Our method relies on pretrained body-part phase networks, which may need to be retrained when the skeleton topology or dataset changes. 
The phase parameterization is tailored to periodic motion components and may be less effective for highly aperiodic, contact-rich, or long-horizon behaviors. 
Addressing these limitations, extending the control space to richer motion attributes, and improving robustness across diverse skeletons and behaviors are promising directions for future work.